\documentclass[preprint]{vgtc}

\graphicspath{{figures/}{pictures/}{images/}{./}}

\usepackage{times}

\usepackage{tabu}
\usepackage{booktabs}
\usepackage{multirow}
\usepackage{lipsum}
\usepackage{mwe}
\usepackage{array}
\newcommand{\PreserveBackslash}[1]{\let\temp=\\#1\let\\=\temp}
\newcolumntype{C}[1]{>{\PreserveBackslash\centering}p{#1}}
\newcolumntype{R}[1]{>{\PreserveBackslash\raggedleft}p{#1}}
\newcolumntype{L}[1]{>{\PreserveBackslash\raggedright}p{#1}}

\usepackage{mathrsfs}
\usepackage{amsmath}
\usepackage{enumitem}
\setitemize[1]{itemsep=0pt, partopsep=0pt, parsep=\parskip, topsep=5pt}
\usepackage{subcaption}
\usepackage{threeparttable}
\usepackage{multirow}
\usepackage{amsfonts}
\usepackage{makecell}

\usepackage[linesnumbered,ruled,vlined]{algorithm2e}

\let\oldthebibliography\thebibliography
\renewcommand{\thebibliography}[1]{%
  \oldthebibliography{#1}%
  \setlength{\itemsep}{0pt}
  \setlength{\parskip}{0pt}
  \setlength{\baselineskip}{9.4pt}
}
\usepackage{mathtools}
\newlength{\myeqskip}  
\AtBeginDocument{%
    \setlength\abovedisplayskip{\myeqskip}%
    \setlength\belowdisplayskip{\myeqskip}%
    \setlength\abovedisplayshortskip{\myeqskip-\baselineskip}%
    \setlength\belowdisplayshortskip{\myeqskip}}

\newif\ifshowrevisions
\newif\ifshowpsy
\showrevisionsfalse
\showpsytrue

\newcommand{\revise}[1]{%
  \ifshowrevisions
    \textcolor{blue}{#1}%
  \else
    #1%
  \fi
}

\onlineid{1295}

\vgtccategory{Research}

\vgtcinsertpkg

\title{Will You Be Aware? Eye Tracking--Based Modeling of Situational Awareness in Augmented Reality}

\author{Zhehan Qu \\ \scriptsize Department of Computer Science \\\scriptsize Duke University %
\and Tianyi Hu \\ \scriptsize \makecell{Department of Electrical and\\ Computer Engineering} \\\scriptsize Duke University
\and Christian Fronk \\ \scriptsize \makecell{Department of Electrical and\\ Computer Engineering} \\\scriptsize Duke University
\and Maria Gorlatova\thanks{e-mail: \{zhehan.qu,tianyi.hu,christian.fronk,maria.gorlatova\}@duke.edu} \\ \scriptsize \makecell{Department of Electrical and\\ Computer Engineering} \\\scriptsize Duke University}

\teaser{
  \centering
  \vspace{-5px}
  \includegraphics[width=\linewidth]{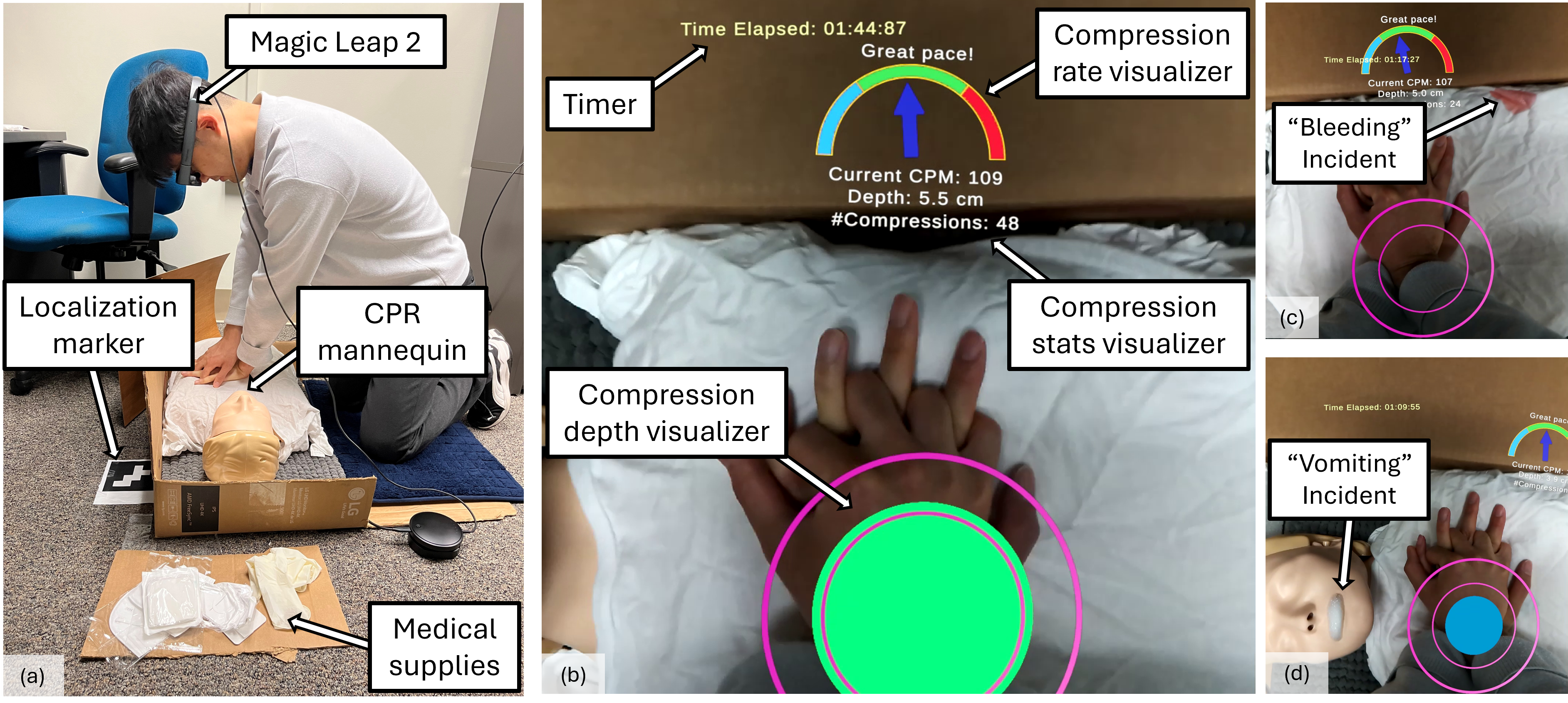}
  \vspace{-16px}
  \caption{A participant performing AR-guided CPR while being evaluated for situational awareness (SA) in response to simulated incidents. (a) A participant wearing the Magic Leap 2 AR headset and performing chest compressions. (b) AR application interface providing real-time CPR guidance. (c) and (d) Simulated ``bleeding'' and ``vomiting'' incidents during the CPR task. Medical supplies were placed to the left of the mannequin, as shown in (a) to allow the user to handle the incidents.}
  \vspace{-2px}
  \label{fig:teaser}
}
\abstract{
    Augmented Reality (AR) systems, while enhancing task performance through real-time guidance, pose risks of inducing cognitive tunneling---a hyperfocus on virtual content that compromises situational awareness (SA) in safety-critical scenarios. This paper investigates SA in AR-guided cardiopulmonary resuscitation (CPR), where responders must balance effective compressions with vigilance to unpredictable hazards (e.g., patient vomiting). We developed an AR app on a Magic Leap 2 that overlays real-time CPR feedback (compression depth and rate) and conducted a user study with simulated unexpected incidents (e.g., bleeding) to evaluate SA, in which SA metrics were collected via observation and questionnaires administered during freeze-probe events. Eye tracking analysis revealed that higher SA levels were associated with greater saccadic amplitude and velocity, and with reduced proportion and frequency of fixations on virtual content. To predict SA, we propose FixGraphPool, a graph neural network that structures gaze events (fixations, saccades) into spatiotemporal graphs, effectively capturing dynamic attentional patterns. Our model achieved 83.0\% accuracy (F1=81.0\%), outperforming feature-based machine learning and state-of-the-art time-series models by leveraging domain knowledge and spatial-temporal information encoded in ET data. These findings demonstrate the potential of eye tracking for SA modeling in AR and highlight its utility in designing AR systems that ensure user safety and situational awareness.
}


\nocopyrightspace

\begin{document}

\maketitle

\section{Introduction}
\vspace{-2px}
Augmented Reality (AR) systems are increasingly deployed as digital overlays to enhance human interaction with physical environments, offering real-time guidance in domains such as industrial maintenance, medical training, and emergency response~\cite{tobiskova2022multimodal,eom2022neurolens,khanal2022virtual,koutitas2021performance}. By superimposing contextually relevant virtual content such as instructional cues~\cite{ceyssens2024art}, navigational aids~\cite{zhang2024exploring}, or procedural prompts~\cite{srinidhi2024xair}, AR optimizes task execution by directing user attention to high-priority information or otherwise invisible elements. However, this strength of AR, which inherently controls visual focus, introduces a critical paradox: while virtual content is designed to streamline task performance, its persistent salience risks monopolizing user attention. Given the finite nature of attentional resources~\cite{wickens2008multiple}, AR overlays that allocate attentional resources in the visual modality to additional virtual elements may diminish the capacity to attend to simultaneous physical stimuli, particularly in dynamic environments where ongoing monitoring is essential.

\vspace{-1px}
Situational awareness (SA), the capacity to perceive, comprehend, and project the state of one's environment~\cite{endsley1995toward}, is indispensable in safety-critical tasks where unanticipated events demand rapid, adaptive responses. While AR has been proposed as a tool to support SA in emergency tasks~\cite{azpiroz2024white} by overlaying real-time procedural guidance or hazard alerts, the technology's emphasis on virtual guidance may inadvertently suppress SA by constraining users' attentional bandwidth. For instance, consider a head-mounted AR system that overlays information on the view of a firefighter in a burning building. As the amount of virtual content, such as biosignals of the rescue team, navigational markers, life supply and device status, increases, the firefighter's ability to monitor the environment and detect rapid life-threatening changes becomes severely compromised. Similar challenges also arise in medical emergencies like cardiopulmonary resuscitation (CPR), where responders must maintain dual attention: executing precise compressions guided by AR while simultaneously monitoring for unpredictable hazards (e.g., patient vomiting or bleeding). While emerging studies~\cite{hou2022comparison,sun2024effectiveness} have demonstrated AR's efficacy in improving CPR task performance, its impact on SA during concurrent real-world disruptions remains underexplored.
AR's virtual content risks disrupting perception by prioritizing guided cues over environmental scanning~\cite{taylor2025co}, while comprehension and projection rely on uninterrupted cognitive bandwidth—resources already strained by multitasking in AR in safety-critical contexts~\cite{hasan2024digital}---all potentially leading to degraded SA and suboptimal outcomes.

\vspace{-1px}
The adverse outcomes of impaired SA highlight the need for effective SA evaluation and modeling in AR. Eye tracking (ET), now standard on most AR headsets, enables assessment of user attention and supports machine learning (ML) models for SA prediction. However, two significant challenges persist. First, accurate measurement of SA in controlled laboratory settings is inherently difficult, as such environments often lack the complexity and unpredictability of real-world, safety-critical scenarios. Traditional assessment methods, such as SAGAT~\cite{endsley1988situation} and SART~\cite{taylor2017situational}, which rely on questionnaires or self-reports, are often subjective and challenging to implement due to the need for freeze-probe techniques. Second, there is a notable scarcity of data capturing user behavior in realistic environments that demand high levels of SA, hindering efforts to model SA using behavioral data. While prior research has predicted suboptimal attention such as mind wandering and distraction using ET data and biosignals~\cite{asish2024classification,bixler2015automatic}, these studies often involved controlled settings where SA is less critical. The lack of data also poses challenges for training robust and generalizable ML models, as user heterogeneity and data scarcity together often lead to severe overfitting and poor generalization. It is therefore crucial to (1) systematically design an AR app and ecologically valid study setup that supports SA evaluation, (2) accurately measure SA and detect related patterns in ET data, and (3) develop robust ML models leveraging patterns in ET that provide reliable SA prediction.

\vspace{-1px}
In this work, we investigate SA of users performing chest compressions, emulating a CPR scenario with a custom AR application developed on Magic Leap 2 for real-time guidance. We developed a novice-friendly app by visualizing compression depth and rate in an in-context, vivid and dynamic manner (guided by American Heart Association (AHA) recommendations~\cite{merchant2020part}), while also incorporating tempo-synchronized drumbeat music~\cite{pellegrino2021value} to facilitate adherence to guideline-recommended chest compression rate. We conducted an IRB-approved user study with 36 participants, in which realistic ``unexpected incidents'' such as the patient bleeding, vomiting or ambulance arriving were activated randomly in the middle of the compression task to test the participants' SA. To evaluate the SA of the participants, we used an ``observation \& questionnaire'' approach, where the observer records the participants' responses to the incidents and the participants also fill out a questionnaire after each incident about their perception, comprehension and decision-making in such situation. Leveraging the capability of the headset, we implement the freeze-probe method by globally dimming the AR display when the incident occurred, and collected ET data for modeling the attention pattern leading to the SA level.

\vspace{-1px}
In addition to identifying that users' SA levels were positively correlated with saccadic amplitude and velocity and negatively correlated with time fixating on virtual content, we further leveraged ET data to model SA levels using ML techniques. Unlike prior studies that primarily focused on predicting activities with consistent and well-defined gaze patterns~\cite{bremer2024predicting,lan2020gazegraph} or internal states using hand-crafted features~\cite{skaramagkas2023esee}, we proposed FixGraphPool, a graph neural network (GNN) based on EdgePool~\cite{diehl2019edge}. FixGraphPool structures gaze events (fixations and saccades) into spatially-and-temporally-connected graphs, effectively capturing the dynamic attentional patterns. The model achieved an accuracy of 83.0\% and an F1 score of 81.0\% in predicting SA levels of unseen users. Our code implementation is publicly available at \url{https://github.com/Duke-I3T-Lab/AR_CPR_SA}. The key contributions of this work are summarized as follows:
\begin{itemize}[itemsep=0pt, topsep=0.3pt, parsep=1pt, leftmargin=0.15in]
    \item We created an \textbf{AR app for CPR} featuring in-context depth visualization (considered ``helpful'' by all but 1 out of 36 users) and conducted a user study with \textbf{realistic ``unexpected'' incidents} during CPR to evaluate user SA.
    \item Through eye tracking analysis, we identified SA to be positively correlated with \textbf{saccadic amplitude and velocity}, and negatively correlated with \textbf{focus on virtual content}, demonstrating the potential negative impact of overwhelming virtual content overlaid on the real world.
    \item We developed a \textbf{graph neural network} based on aggregated gaze events and spatial-temporal connections between them to predict SA level of unseen users, outperforming feature-based classic ML models and SOTA time-series models.
\end{itemize}

\vspace{-6px}
\section{Related Work}
\label{sec: related}
\vspace{-4px}
\noindent\textbf{SA and its evaluation in AR}. 
SA is defined as the perception of elements in the environment, the comprehension of their significance, and the projection of their future states~\cite{munir2022situational}. The widely referenced \textit{Endsley's Cognitive Model}~\cite{endsley1995toward} delineates three hierarchical levels of SA as follows:
\begin{itemize}[itemsep=0pt, topsep=0.2pt, parsep=0pt, leftmargin=0.1in]
  \item \textit{Perception} (Level 1): the detection and recognition of the status, attributes, and dynamics of relevant elements in the environment.
  \item \textit{Comprehension} (Level 2): the integration of Level 1 elements into a coherent understanding of the environment, enabling the interpretation of the significance of objects and events.
  \item \textit{Projection} (Level 3): the ability to anticipate the future states and actions of elements in the environment, based on Levels 1 and 2.
\end{itemize}
SA is regarded as a cornerstone of effective decision-making in critical domains such as aviation, healthcare, and emergency response~\cite{endsley1999situation,sapateiro2009emergency,kedia2022technologies}. However, in the context of AR, comprehensive evaluations of SA and its components remain limited~\cite{woodward2022analytic}, with few works targeting industry~\cite{truong2023study} and aviation~\cite{pan2025situational} scenarios. Other works have primarily focused on related suboptimal attentional states, such as distraction~\cite{kim2022assessing,qu2024looking} and inattentional blindness~\cite{dixon2014inattentional,syiem2021impact}, without explicitly linking (if suitable) these phenomena to a broader SA framework. This study seeks to address this gap by developing a robust scenario for SA evaluation in AR and leveraging ET data to model SA, thereby advancing the understanding of SA in AR environments.

\noindent\textbf{AR in emergency response}. AR has been extensively explored in emergency response, particularly within the domain of emergency medicine~\cite{harari2025applications}. For example, Siebert et al.~\cite{siebert2017adherence} adapted the AHA Pediatric Advanced Life Support guidelines for AR glasses, demonstrating that AR users exhibited reduced errors and deviations in defibrillation doses during CPR. Similarly, an AR training module for operating AmBus significantly decreased task completion times~\cite{koutitas2021performance}; the use of smart glasses for patient assessment in mass-casualty incidents improved both accuracy and speed~\cite{apiratwarakul2022smart}. Integrated systems such as EMSAssist~\cite{jin2023emsassist} and CognitiveEMS~\cite{weerasinghe2024real} can provide emergency responders with real-time information and decision-making support. Beyond emergency medicine, AR has shown potential in enhancing users' knowledge of the environment in scenarios such as building evacuation~\cite{sharma2019emergency}, firefighting~\cite{collington2018pervasive}, and disaster management~\cite{khanal2022virtual}. While the underlying thought behind these applications is to improve SA by augmenting the user's perception, little has been done to understand the impact of AR on higher levels of SA, particularly in the presence of unexpected but critical events. This work seeks to investigate the nuanced effects of AR on user SA in such complex and dynamic scenarios.

\noindent\textbf{Eye tracking--based modeling of user context}. 
A substantial body of research has explored user context sensing through ET, addressing both internal states—such as emotion, mind wandering, and attentional tunneling~\cite{skaramagkas2023esee,bixler2015automatic,kortschot2020classification}—and external states such as activity recognition~\cite{wang2020capture,lan2020gazegraph}. With the increasing integration of eye-tracking cameras in AR and VR headsets, recent studies have leveraged this technology to sense user context in immersive environments, focusing on activities and locomotion~\cite{scargill2022demo, bremer2024predicting}, attentional states~\cite{qu2024looking}, and cybersickness~\cite{jeong2022eyes}. 
Modeling approaches in this domain can be broadly categorized into classical ML models employing hand-crafted features~\cite{skaramagkas2023esee,sharma2022student} and deep learning (DL) models that operate directly on raw data, such as gaze-annotated images~\cite{vortmann2021imaging}, time series~\cite{bremer2024predicting,qu2024looking}, or graphs~\cite{lan2020gazegraph,fan2019understanding}. While DL models often demonstrate superior performance compared to traditional methods, they typically require well-defined, user-agnostic tasks characterized by consistent gaze patterns and easily processable gaze data. 
However, these conditions are rarely met in real-world scenarios, especially when users interact with the 3D environment around them. To address this limitation, we propose a GNN model that leverages extracted gaze events to robustly predict situational awareness levels in the dynamic and unpredictable context of CPR tasks.

\vspace{-4px}
\section{Task for Evaluation of SA in AR: CPR with Random Incidents}
\vspace{-2.5px}
\label{sec: app}
\subsection{Motivation}
\vspace{-3px}
To effectively model SA in AR, the task must enable objective, reliable, and meaningful SA measurement. Comparing SA evaluation tasks (e.g., pilot training~\cite{wickens2009attentional}) with typical AR tasks, we identify three key distinctions:
\begin{itemize}[itemsep=0pt, topsep=0.1pt, parsep=0pt, leftmargin=0.1in]
  \item \textit{Task scope}: SA evaluation tasks require comprehensive situational monitoring and decision-making, while typical AR tasks often focus on a single objective with limited elements.
  \item \textit{Requirement for environmental awareness}: SA is relevant when tasks necessitate awareness of the broader setting, unlike many AR tasks where attention is confined to predefined objectives.
  \item \textit{Environmental dynamics}: SA is meaningful in dynamic, event-driven contexts, while AR tasks are often static and predictable.
\end{itemize}
We selected CPR for this study due to its alignment with the requirements for SA evaluation. The task inherently demands continuous awareness of the patient's overall condition and the surrounding environment, where lapses in SA can have life-threatening consequences. CPR can also incorporate realistic, dynamic incidents---identified in consultation with a certified CPR instructor---that can be effectively simulated in a controlled laboratory setting. Furthermore, CPR provides objective performance metrics, such as compression depth and rate, which can guide the design of the AR application. This application is envisioned for future deployment in real-world emergencies to assist both trained or untrained individuals. The following sections detail the AR application and the incidents developed for SA evaluation.

\vspace{-5px}
\subsection{AR App for CPR Guidance}
\vspace{-3px}
\noindent\textbf{App design and features.} The AR application we developed incorporates four primary visual components: a compression depth visualizer, a compression rate visualizer, a compression stats visualizer and a timer, as illustrated in Fig.~\ref{fig:teaser}(b). Targeting novice users, the major difference between our app and existing AR CPR apps~\cite{leary2020pilot,hou2022comparison} developed for training purposes is the in-context compression depth visualizer placed right above the user's wrist, consisting of two concentric rings and a dynamic circular depth indicator. The indicator's size and color are adjusted based on the depth of the chest compressions. Specifically, the color transitions along a gradient from blue at shallow compression depth, to green for the recommended depth range of 5--6 cm~\cite{merchant2020part}, to red for excessive depth (though it was often seen as yellow since the maximum depth was seldom reached). This design was inspired by an AR force visualizer used in robotic ultrasound applications~\cite{song2024optimizing}, adhering to the blue-green-red color-coding scheme proposed by Ochitwa et al.~\cite{ochitwa2024msk}. The two concentric rings, representing the boundaries of the recommended depth range, have diameters of 9~cm and 13~cm, respectively, while the depth indicator circle can expand up to a maximum diameter of 16~cm.

The compression rate visualizer, placed on the top-right of the compression position, is designed as a speed gauge with a diameter of 10~cm, using a blue arrow to indicate the current compression rate. The gauge employs the same color-coded scheme as the one used in Laerdal Medical's QCPR app: blue for rates below the recommended range, green for rates in the optimal range of 100--120 compressions per minute, and red for rates exceeding the recommended range. Two text boxes accompany the gauge: one provides real-time feedback on the compression rate (e.g., ``Speed up!'', ``Slow down!'', or ``Good pace!''), while the other (the stats visualizer) displays the current compression rate and depth.

The timer, fixed in the top-left corner of the user's field of view (avoiding visual clutter), displays the total elapsed time since the start of the task, emphasizing the criticality of time in life-saving scenarios~\cite{wickens2022applied}. Additionally, for novices to conduct effective CPR, we followed the common usage of music in CPR training~\cite{pellegrino2021value} and incorporated a tempo-synchronized drumbeat at 108 beats per minute to help users adhere to the recommended compression rate. We developed this version of the app after conducting a pilot study with 10 participants, who provided feedback on the app's usability and effectiveness. The final app design was informed by a clinical nurse specialist at our institution, though we acknowledge limited design iteration may impact our findings, as discussed in Sec.~\ref{sec: discussion}.

\noindent\textbf{Implementation details.} The app was developed using Unity 2022.3.42f1 and deployed on a Magic Leap 2. The compression stats were obtained via a Google Pixel 3 XL running the QCPR app developed by Laerdal Medical, that was Bluetooth-connected to a Laerdal Resusci Anne QCPR Mannequin. Connection between the phone and the app was established via a local Wi-Fi network.

\vspace{-5px}
\subsection{Random Incidents for SA Evaluation}
\vspace{-3px}
\noindent\textbf{Incidents identification.} In consultation with a certified CPR instructor and trainer, we identified two realistic incidents relevant to CPR that could be simulated in a controlled laboratory setting. These incidents and methods to handle them are as follows:
\begin{itemize}[itemsep=0pt, topsep=0.2pt, parsep=0pt, leftmargin=0.1in]
  \item \textit{Bleeding or open wounds}: Bleeding should be managed using gauze or bandages while ensuring uninterrupted CPR.
  \item \textit{Vomiting or regurgitation}: Common during CPR particularly if the victim has a full stomach. One should turn the patient to their side to prevent aspiration, clear the airway with tools such as gauze or a gloved hand, and return the patient to a supine position to resume CPR.
\end{itemize}
Additionally, we developed a third incident in which a virtual \textit{Ambulance} arrives, simulating the arrival of professional help. The ambulance would approach with a siren sound from the front-left of the user. We expect the user to locate the ambulance and stop CPR to hand over the patient to the ambulance crew. 

\begin{figure}[t]
	\centering
  \subcaptionbox{Interior of the mannequin}
	{\includegraphics[height=4.2cm]{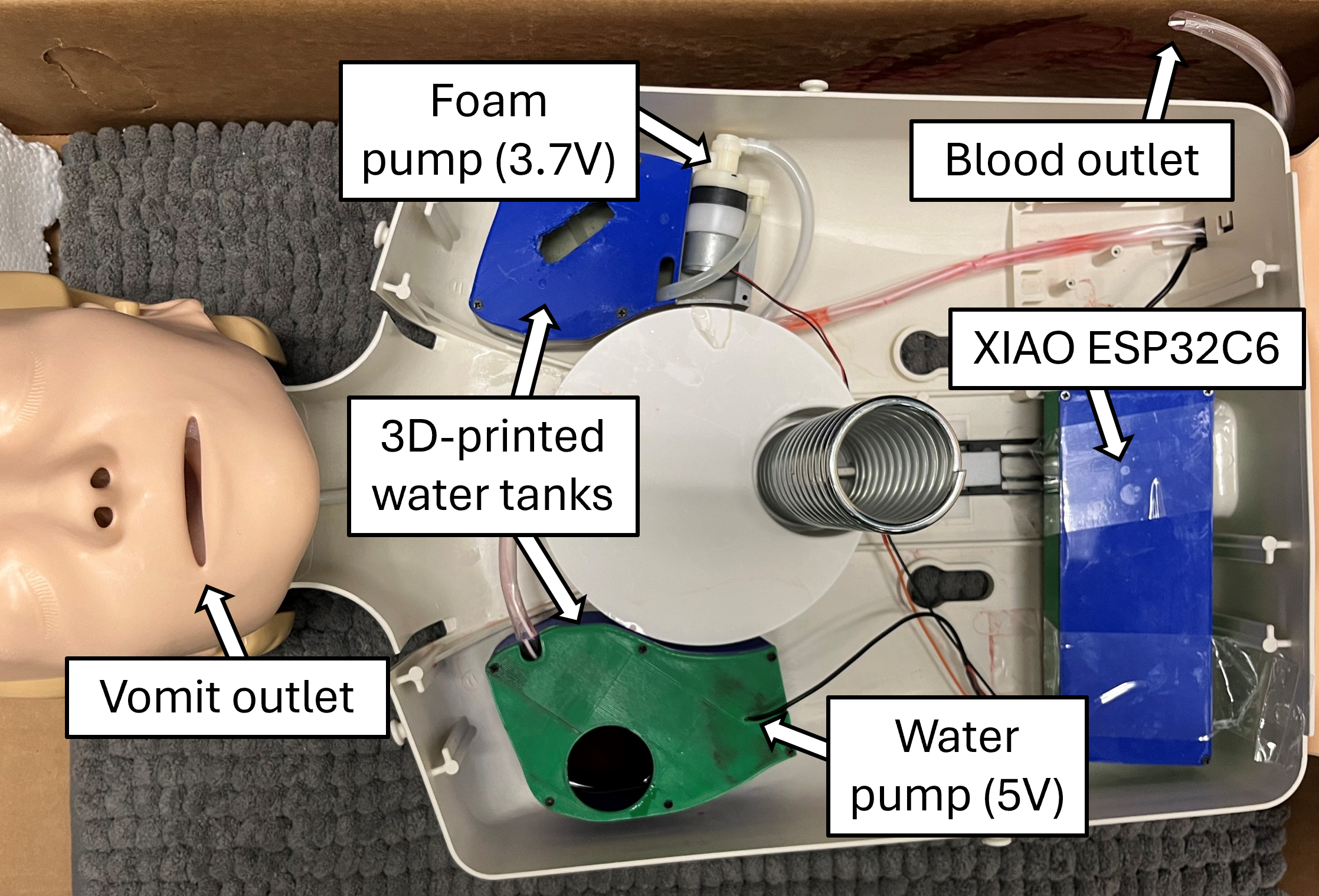}}
    \hspace{0.2mm}
	\subcaptionbox{Android app}
	{\includegraphics[height=4.2cm]{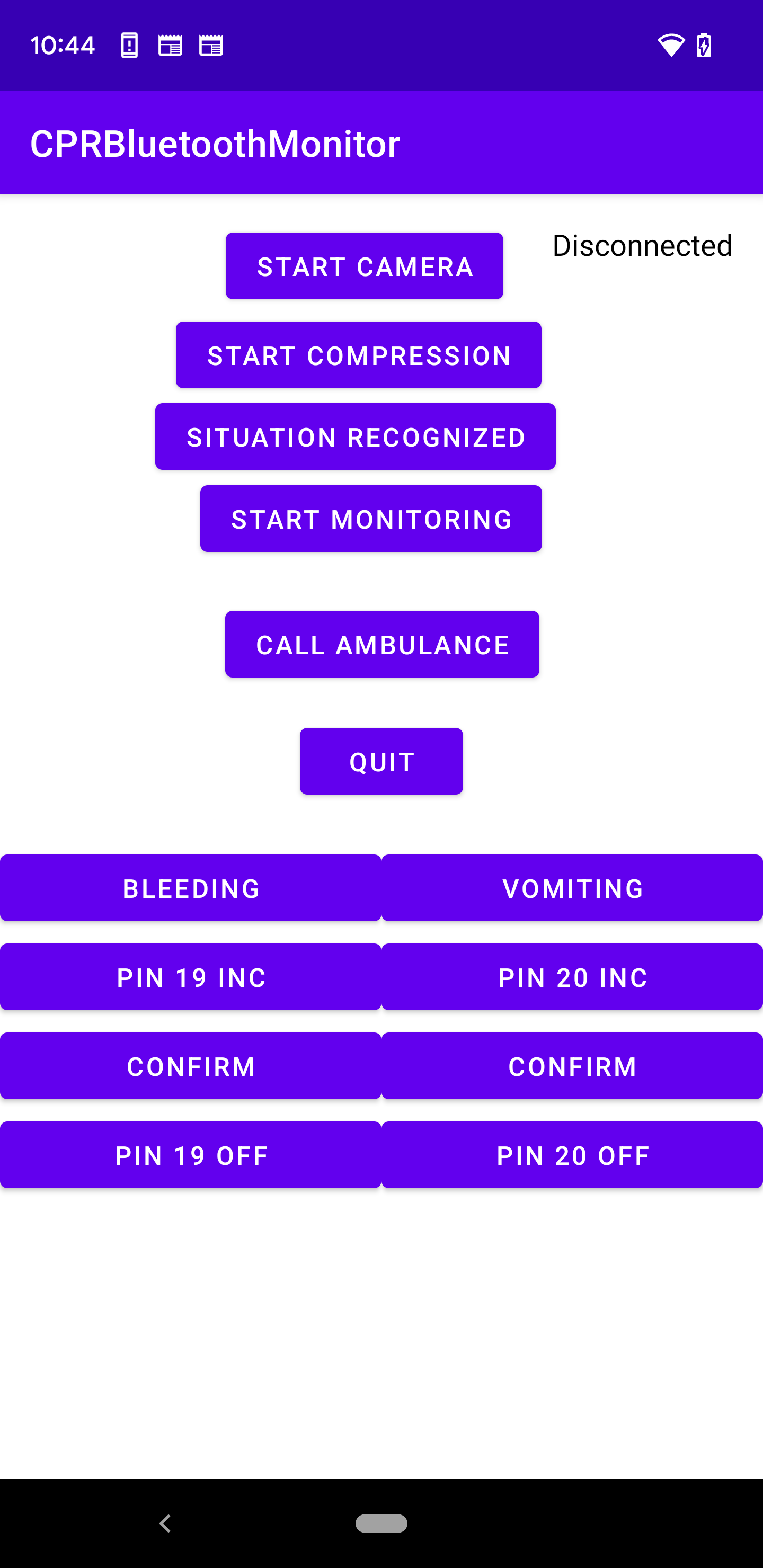}}	
    \vspace{-8px}
	\caption{Artifacts used for generating the incidents. (a) The interior of the mannequin equipped with water tanks, tubes and microcontroller. (b) Interface of the Android app used to control the incidents. The app also served as a controller for activating video capture and session start command in the AR app. }
    
  \label{fig: incidents-implementation}
  \vspace{-20px}
\end{figure}

\noindent\textbf{Implementation.} The bleeding (B) and vomiting (V) incidents were implemented using a custom setup, as illustrated in Fig.~\ref{fig: incidents-implementation}. Two 3D-printed tanks were embedded within the mannequin, each designed to release liquid upon activation. The ``blood'' tank, filled with water dyed red, was connected to a water pump, while the ``vomit'' tank, containing a mixture of water and soap to simulate foam, was connected to a foam pump. These artifacts were placed in a way that does not affect the mannequin's functionality. The blood was released from the mannequin's left lateral abdominal region, and the vomit from its mouth, with both release points positioned 21 cm away from the compression point to maintain comparable detection difficulty. During the study, a white shirt was placed on the mannequin to conceal the tubing for the blood simulation. Images of B and V incidents are shown in Fig.~\ref{fig:teaser}(c) and Fig.~\ref{fig:teaser}(d).

The ambulance incident was implemented as a virtual ambulance approaching the user from a distance of 95~m at 6~m/s, stopping 5~m away. The simulation included a siren sound with a Doppler effect to enhance realism. Real incident activation was controlled via an ESP32C6 microcontroller, which was integrated with the same Android application used for capturing compression statistics. The android app sends commands to the microcontroller via HTTP requests to activate the incidents, and directly sends messages to the Magic Leap 2 app via TCP to trigger the ambulance incident.

\vspace{-3px}
\section{User Study for Situational Awareness Modeling in AR-Guided CPR}
\vspace{-2px}
\label{sec: study}
With the aforementioned setup, we conducted a user study to evaluate the impact of AR of user SA in CPR. The purpose of the study was to collect ET data and user SA labels to conduct ET-based modeling of SA, as well as training and testing ML models for SA prediction. The study was approved by Duke University Campus Institutional Review Board.
\vspace{-5px}
\subsection{Participants and Environment}
\vspace{-3px}
We recruited 36 participants for the study and excluded 6 due to data collection issues (e.g., ET or rendering failures). The final cohort of 30 participants had an average age of 23.5 years, comprising 19 males, 9 females, and 2 identifying as ``other.'' 11 of them had never worn an AR headset before, while 5 wear one at least once a week. Recruitment did not target a specific level of CPR expertise, resulting in a cohort with diverse experience: 9 participants had no prior hands-on CPR experience, 16 had basic awareness or informal training without certification, 2 were certified CPR instructors, and 3 held Basic Life Support certification.
The cohort's high level of CPR expertise is likely attributable to recruitment in a locale with a significant healthcare presence.

The study was conducted in a quiet room with controlled lighting conditions (shown in Fig.~\ref{fig:teaser} (a)). The mannequin was placed in a cardboard box that was taped to the floor, with the mannequin fixed in the cardboard box using styrofoam. An April Tag was taped behind the cardboard box on the floor to ensure consistent placement of the AR compression rate visualizer. Rugs were placed beneath the mannequin to prevent water from leaking out of the box and to provide a comfortable surface for the participants. A white shirt was placed on the mannequin to simulate a real patient and also to hide the tube for delivering ``blood''. Two black pieces of tape were used to mark the correct position for compression. A set of tools including gauze, masks and medical gloves were placed to the left of the mannequin, so that the user can easily access them if needed.

\vspace{-5.5px}
\subsection{Study Procedure}
\vspace{-3.5px}
Upon arrival, participants signed a consent form. They then completed a pre-study survey, including the \textit{ATTC} questionnaire~\cite{derryberry2002anxiety}, which evaluates attention focus (\textit{ATTCF}) and shifting (\textit{ATTCS}) abilities, combined into a single attention control score. Participants also completed two additional cognitive tests: the Flanker Squared task~\cite{burgoyne2023nature} for attention control (score denoted as \textit{ACS}) and the shortened Symmetry Span task~\cite{oswald2015development} for working memory capacity (score denoted as \textit{WMCS}). These scores were later analyzed for correlations with SA levels observed during the study.

Participants were introduced to CPR through a video and the AR app through a hands-on tutorial. After eye calibration, they practiced using the app and performed standard compressions on the mannequin, during which baseline ET data, including pupil size, were recorded. Participants then reviewed an introductory description of the study, which explicitly instructed them to ``instead of thinking of it as training, try to perform chest compressions while monitoring the victim's condition and surroundings, using your own judgment to respond to unusual events.'' Note that no artificial respiration was required in the study, and we informed the users to keep the compression going before they notice anything unusual. 

\vspace{-3px}
\begin{table}[h]
  \vspace{-5px}
  \caption{Perception and comprehension-level questions asked for SA evaluation.}
  \vspace{-8px}
  \label{tab: questions}
  \setlength{\tabcolsep}{3pt}
  \begin{tabular}{@{}ccl@{}}
  \toprule
  Incident                                       & Level & Question                                                                                       \\ \midrule
  \multirow{2}{*}{Bleeding}                      & Perc. & Where on the shirt got wet?                                                                    \\
                                                 & Comp. & What did the shirt getting wet indicate?                                             \\\hline
  \multirow{2}{*}{Vomiting}                      & Perc. & What was the color of the liquid?                            \\
                                                 & Comp. & What happened on the patient?                                                  \\ \hline
  \multirow{2}{*}{Ambulance} & Perc. & Where did the vehicle come from? \\
                         & Comp. & What was the vehicle and what did it mean?                \\ \bottomrule
  \end{tabular}
  \vspace{-10px}
\end{table}

After confirming their understanding, participants completed three randomized trials, one for each incident type. In each trial, participants performed compressions for up to one minute, with the incident randomly triggered between 30 and 40 seconds. The observer recorded participants' responses. If the user initiated any action or if no response occurred within 5 seconds (10 seconds for the ambulance trial), the trial ended using Magic Leap 2's ``global dimming'' feature to block the user's view. Participants then answered SA questions designed based on SAGAT~\cite{endsley1988situation}, as shown in Table~\ref{tab: questions}. Instead of projection-level questions which are often vague and difficult to validate~\cite{truong2023study}, participants were asked a \textit{decision-making}-level question: ``Given your comprehension of the current situation, what do you plan to do?'' Their responses were compared with the observer's records to help infer SA levels. We developed this approach based on our pilot study and further explanation is provided in the next section. Additionally, participants completed the \textit{SART} questionnaire~\cite{taylor2017situational} after each trial, which evaluates SA across three dimensions: understanding, attentional \textit{demand}, and attentional \textit{supply}. The overall SA score was calculated as:
\vspace{-2px}
\begin{equation}
  \text{SA}_{\text{SART}} = \text{Understanding} - (\text{Demand} - \text{Supply}).
  \vspace{-2px}
\end{equation}
This subjective measure of SA served as a secondary evaluation, complementing the more objective SA levels determined through observations and participant responses.

After completing all trials, participants provided open-ended feedback on the app. They were compensated with snacks and souvenirs. Each session lasted approximately one hour.

\vspace{-6px}
\subsection{Measurements and Data Collection}
\vspace{-3px}
\noindent\textbf{Exclusion of the ambulance incident.} Before discussing the measurement details, it is important to note that we excluded the ambulance trial from the analysis. This decision was based on the observation that only 1 out of 36 participants attempted to locate the ambulance, with the majority interpreting the sound of the siren as the conclusion of the trial, making it impossible to properly assign SA labels. Given that the ambulance trial involves no real-world stimuli and therefore did not prepare the user for subsequent incidents, we excluded this trial to exclusively focus on the two realistic and contextually relevant incidents (B and V), which are more aligned with the objectives of SA evaluation. Of the 30 analyzed participants, the ambulance incident was presented first, second, and last for 9, 10, and 11 participants, respectively. After exclusion, 13 participants experienced the bleeding trial before the vomiting trial. Further analysis on the exclusion is provided in Sec.~\ref{sec: population-statistics}.

\noindent\textbf{Measurement of SA.} User SA was categorized as \textit{good} or \textit{poor} following the procedure in Algorithm 1, which applied a flexible assessment to decision-making. While most \textit{poor} SA instances stemmed from failed perception, a \textit{poor} label also resulted from subsequent failures such as: inadequate comprehension despite perception (e.g., \textit{P5} saw ``blood'' but stated, ``Somehow I did not know what happened,'' and continued compressions); unreasonable decisions (e.g., \textit{P25} planned to compress ``faster but less hard''); or actions inconsistent with stated decisions (e.g., \textit{P14} intended to ``wipe the liquid and keep on'' but did not). This flexible approach to evaluating decision-making was adopted to accommodate diverse participant experience and medical knowledge; for instance, while a medical professional might argue that there are only specific correct ways to handle the incidents, a layperson stating, ``I planned to wait for professional help,'' and therefore stopping compressions, is also a reasonable response that should not be labeled as \textit{poor} SA. Accordingly, 5 out of 60 trials received a \textit{poor} SA label due to failures at the comprehension or decision-making levels, despite successful initial perception of the incident.

\vspace{-10px}
\begin{algorithm}[h]
  \caption{Labeling procedure for SA}
  \SetAlgoNlRelativeSize{-1} 
  \SetAlgoVlined 
  \KwIn{Answers to SA questions, Observed behavior}
  \KwOut{SA label (\textit{good} or \textit{poor})}
  \If{Perception-level answer is incorrect or not perceived}{
      \Return \textit{poor}
  }
  \If{Comprehension-level answer is incorrect or unanswered}{
      \Return \textit{poor}
  }
  \If{Decision-making answer lacks common sense or does not match observed behavior}{
      \Return \textit{poor}
  }
  \Return \textit{good}
  \label{alg:sa_labeling}
\end{algorithm}
\vspace{-12px}

\noindent\textbf{Data collection.} Using the Magic Leap OpenXR Eye Tracker Feature, we collected eye tracking data at 60Hz, including pupil diameter, cyclopean gaze direction, cyclopean eye center position, whether a blink was identified and whether the gaze targeted a virtual object at each timestamp. Ego-centric videos were used to timestamp user reactions to incidents.

\begin{figure*}[t]
	\centering
	\includegraphics[width=\linewidth]{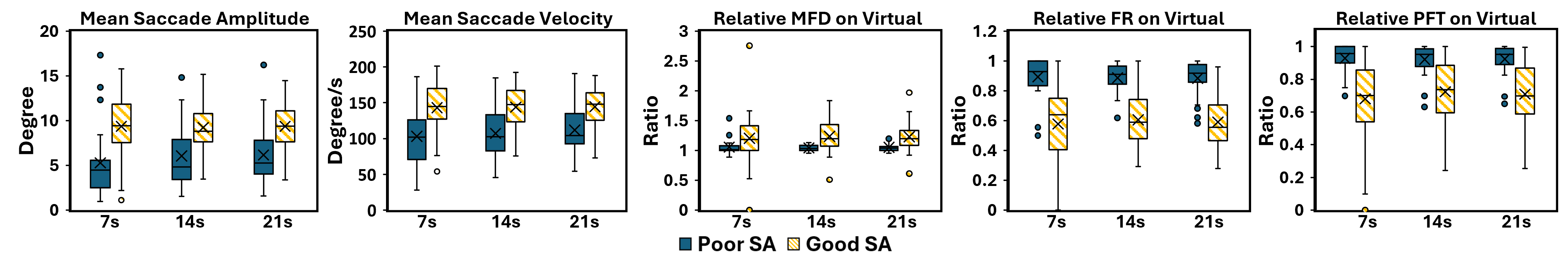}
  \vspace{-6.5mm}
	\caption{Metrics of gaze events in different time windows before SA evaluation, comparing \textit{good} and \textit{poor} SA levels. Note that the ROI-related metrics are computed as ratios to the overall metric. Best viewed in color.}
  \vspace{-20px}
  \label{fig: gaze-results}
\end{figure*}

\vspace{-5px}
\section{Study Results}
\vspace{-2px}

\label{sec: analysis}
This section presents the study results, focusing on the relationship between self-reported measures, ET data, and SA levels. We consider that maintaining good SA in this task requires (1) sufficient working memory to manage CPR and environmental monitoring simultaneously, (2) broad attention and efficient visual processing to quickly detect and interpret incidents, and (3) adequate cognitive resources allocated to the environment. Below, we first analyze SA's relationship with subjective data, followed by ET metrics.

\vspace{-3px}
\subsection{Population Statistics}
\label{sec: population-statistics}
\vspace{-2px}
For the 30 users analyzed, the cognitive assessment scores are shown in Table~\ref{tab:cognitive-scores}. The \textit{WMCS} was measured using the edit-distance scoring method~\cite{gonthier2023easy}.
\begin{table}[t]
  \centering
  \caption{Participant cognitive assessment scores (n=30).}
  \vspace{-8px}
  \label{tab:cognitive-scores}
  \setlength{\tabcolsep}{3pt}
  \begin{tabular}{lccc}
    \toprule
    \textbf{Measure} & \textbf{Mean} & \textbf{SD} & \textbf{Range} \\
    \midrule
    Attention Control Scale (\textit{ATTC}) & 55.4 & 7.1 & 43--70 \\
    Attention Focus (\textit{ATTCF}) & 25.2 & 3.1 & 20--31 \\
    Attention Shifting (\textit{ATTCS}) & 30.2 & 5.0 & 22--40 \\
    Attention Control Score (\textit{ACS}) & 40.3 & 9.7 & 20--56 \\
    Working Memory Capacity Score (\textit{WMCS}) & 0.75 & 0.13 & 0.48--1 \\
    \bottomrule
  \end{tabular}
  \vspace{-20px}
\end{table}

For SA measurements, out of the 60 incidents (30 B and 30 V), users in 21 B trials and 15 V trials were labeled as having \textit{good} SA, while the remaining 24 trials were labeled as \textit{poor} SA. The \textit{SART} questionnaire yielded an average SA score of 16.1 for the B incident and 13.7 for the V incident. This difference, found significant by a t-test ($p<.01$), suggests that the V incident induced higher attentional demand and was more challenging to comprehend, aligning with our SA labeling results. Additionally, a Pearson correlation of 0.26 ($p=.04$) was observed between the \textit{SART} score and the SA labels across all trials, indicating a moderate relationship between subjective SA scores and the labels we assigned. We additionally tested an alternative labeling method in which one must correctly answer all three levels of questions to be labeled as \textit{good} SA. Labels obtained using this method were found to be not significantly correlated with the \textit{SART} score ($r=0.15, p=.26$), suggesting that the labeling method we adopted is more suitable for this study.

\vspace{-1px}
To further examine the factors influencing SA, we fit a linear mixed-effects model to the SA labels, with incident type and trial order as fixed effects and user ID as a random effect. Random intercepts for user ID accounted for 51\% of the variance in SA (Var = 0.109, SD = 0.33), with residual variance at 0.106 (SD = 0.33). Trial order was found to have a significant positive effect on SA ($\beta=0.31, SE=0.09, t(28.0)=3.68, p<.001$), suggesting that participants improved their SA over successive trials. Such improvement was also observed if we kept the original order within all 3 trials in the model ($\beta=0.24, SE=0.06, t(35.3)=4.25, p<.001$). However, the effect of incident type was not significant in either model, supporting the decision to merge data from both incidents for training and testing in the SA prediction model.
\vspace{-1px}

Finally, we analyzed the correlation between SA labels and pre-study measurements. Surprisingly, no significant correlation was found between SA and any pre-study metrics when considering all trials combined. However, when analyzing the V trials separately, a moderate negative correlation was observed between \textit{ATTCS} and SA labels ($r=-0.51, p<.01$). Further evidence from ET metrics, discussed next, provides more context for this finding.

\vspace{-6px}
\subsection{ET-Based Analysis}
\vspace{-4px}
We consider user gaze behavior and pupil size as indicators of SA, as they reflect the users' cognitive state and attentional focus~\cite{dehais2020neuroergonomics}. Given our labels for each trial as binary representations of SA, this section examines the differences in eye-tracking metrics between good and poor SA levels, controlling for the trial order.

\vspace{-6px}
\subsubsection{ET Data Preprocessing}
\vspace{-4px}
\label{sec: gaze-processing}
Our pipeline began by transforming gaze data into the marker's coordinate system to improve filter accuracy, as recommended by Manakov et al.~\cite{manakhov2024filtering}. For event classification, we used an Identification by Velocity Threshold~\cite{salvucci2000identifying} algorithm modified with a dynamic, dual-velocity threshold, which distinguishes between conditions with and without significant head movement. Following established guidelines~\cite{olsen2012tobii}, we applied a 30 deg/s velocity threshold for periods of little to no head movement. \revise{When head motion velocity surpassed 16.8~cm/s (the minimum of all users when conducting compressions), the threshold was raised to 100~deg/s. We adopted this higher threshold because, from a velocity-based classification perspective, the compensatory eye movements generated by vestibulo-ocular reflex fall into the same ``slow-phase'' category as smooth pursuits, which are typically identified at 100~deg/s~\cite{luna2008development,meyer1985upper}}. Both events were merged before further processing, aligning with the common practice of treating fixations and smooth pursuits as a unified measure of visual attention. The remaining gaze events (excluding blinks) were classified as saccades. The mean saccade duration across participants under the adopted thresholds was 73.9~ms, consistent with typical saccade durations~\cite{yarrow2004consistent}.

Pupil size data were normalized per participant using baseline measurements from the practice trial. Throughout the paper, ``pupil dilation'' refers to the z-scores of normalized pupil size data.

\vspace{-4px}
\subsubsection{Metrics}
\vspace{-3px}
\label{sec: gaze-metrics}
With a focus on selecting ET metrics related with user attention and SA, we computed the following metrics for each trial:
\begin{itemize}[itemsep=0pt, topsep=0.1pt, parsep=0.3pt, leftmargin=0.1in]
  \item \textit{Fixation-related}: Fixation rate (FR, number of fixations per second), mean fixation duration (MFD) and proportion of fixated time (PFT). Higher FR is typically related with higher perceptual load, while higher MFD and PFT relate to higher focus and cognitive load~\cite{liu2022assessing}\revise{.}
  \item \textit{Saccade-related}: Mean saccade amplitude (MSA), mean saccade velocity (MSV) and mean peak saccade velocity (MPSV). We consider saccade features to be important in SA modeling, not only because they encode the breadth of visual attention, but also because saccades were found to be related with higher-level cognitive processes such as confidence in decision-making~\cite{seideman2018saccade} and the encoding of spatial information~\cite{mostofi2020spatiotemporal}.
  \item \textit{ROI-related}: Proportion of fixated time, mean fixation duration, and fixation rate on virtual content (VPFT, VMFD, VFR) \textit{divided by} those of all fixations, indicating the relative level of cognitive focus on virtual content.
  \item \textit{Others}: Blink rate (BR) and mean pupil dilation (MPD), linking to cognitive load~\cite{ledger2013effect,piquado2010pupillometry}\revise{.}
\end{itemize}

We consider eye movements to be indicative of SA within a time window preceding the incident. Based on pilot study observations, user behavior remained consistent for approximately 20 seconds prior to incidents. Therefore, we analyze metrics extracted from three time windows: 7s, 14s, and 21s before the incident. Below, we highlight the stability of user patterns during this period and identify metrics significantly correlated with SA levels.

\vspace{-5px}
\subsubsection{Analysis of ET Metrics}
\vspace{-3px}
\noindent\textbf{Relationship with SA.} We analyzed the association between ET metrics and SA using linear mixed-effects models, incorporating SA and trial order as fixed effects and user ID as a random effect. Table~\ref{tab: stats} summarizes significant results after applying Bonferroni correction ($p<.017$).
\vspace{-1px}
The results were consistent across different window lengths, although some fluctuations in significance were observed, particularly in the 7-second window. These fluctuations may be attributed to the smaller number of events captured within this shorter time frame compared to the other two windows. No significant effect of trial order on gaze metrics was found when controlling for SA. Saccadic metrics (MSA, MSV, MPSV) showed significant positive associations with SA, indicating that participants with better SA exhibited faster and longer saccades. This suggests that efficient spatial encoding may have facilitated quicker comprehension of incidents and improved decision-making.

Additionally, VMFD was positively correlated with SA, while VFR and VPFT were negatively correlated. These findings suggest that participants with good SA allocated more cognitive resources per fixation to virtual content but reduced the overall number and duration of such fixations, thereby dedicating more attention to the real environment. This behavior contrasts with cognitive tunneling, often linked to poor SA. Fig.~\ref{fig: gaze-results} provides a visual comparison of these metrics between good and poor SA levels.

\begin{table}[t]
  \centering
  \caption{Effect estimates of SA on ET metrics over different time preceding the incident. Insignificant results are shown in \textit{italics}.}
  \vspace{-2px}
  \label{tab: stats}
  \begin{tabular}{@{}c|cc|cc|cc@{}}
  \toprule
  & \multicolumn{6}{c}{Window length}   \\  
  & \multicolumn{2}{c}{7s}  \vline & \multicolumn{2}{c}{14s}  \vline& \multicolumn{2}{c}{21s} \\
  \multirow{-3}{*}{Metric} & $\beta$ & $p$ & $\beta$ & $p$ & $\beta$ & $p$  \\ \midrule
  MSA & 3.11 & .004 & 2.61 & .005 & 2.79 & .001 \\ 
     MSV  & \textit{22.95} & \textit{.04} & 26.75 & .005 & 24.99 & .005\\ 
     MPSV  & 42.55 & .01 & 43.25 & .003 & 43.97 & .002 \\

     VMFD & \textit{0.152} & \textit{.10} & 0.160 & .007 & 0.138 & .014 \\ 
     VFR & -0.293 & $<$.001 & -0.240  &$<$.001 & -0.256 & $<$.001 \\ 
     VPFT  & -0.239 & $<$.001 & -0.184 & $<$.001 & -0.190 & $<$.001 \\
    \bottomrule
  \end{tabular}
  \vspace{-20px}
  \end{table}

Notably, MFD also showed a near-significant negative relationship with SA ($\beta=-0.22, p=.022$ for the 7s window; $\beta=-0.13, p=.09$ for the 14s window; $\beta=-0.14, p=.037$ for the 21s window). While not included in the main results due to correction and variability across windows, this trend suggested that participants with good SA may have exhibited shorter fixation durations, potentially reflecting more efficient visual processing.

\noindent\textbf{Relation with user characteristics}. To further investigate the observed negative correlation between \textit{ATTCS} and SA labels, we examined the relationship between \textit{ATTCS} and ET metrics within the 21-second window preceding the incidents. The analysis revealed moderate negative correlations between \textit{ATTCS} and MSV ($r=-0.40, p=.03$), MPSV ($r=-0.43, p=.02$), and MSA ($r=-0.42, p=.02$). While \textit{ATTCS} measures an individual's self-reported ability to shift attention across multiple tasks, these findings suggest that participants with higher \textit{ATTCS} scores may exhibit less dynamic visual processing. This observation underscores the need for further investigation, as elaborated in the Sec.~\ref{sec: discussion}. 

Additionally, \textit{WMCS} was found to be negatively correlated with FR ($r=-0.32, p=.01$) and positively correlated with MFD ($r=.34, p=.004$) and VMFD ($r=.24, p=.06$). These results suggest that individuals with higher WMC may have allocated more cognitive resources to the task particularly in interpreting and utilizing the virtual content provided, as higher fixation duration and lower fixation rate were shown to be related with high cognitive load~\cite{liu2022assessing,meghanathan2015fixation}. However, the correlation between \textit{WMCS} and SA labels was not statistically significant. Future work will aim to further investigate the nuanced relationship between WMC and SA.

\begin{figure*}[t]
	\centering
	\subcaptionbox{Example graph constructed from window data}
	{\includegraphics[height=6.4cm]{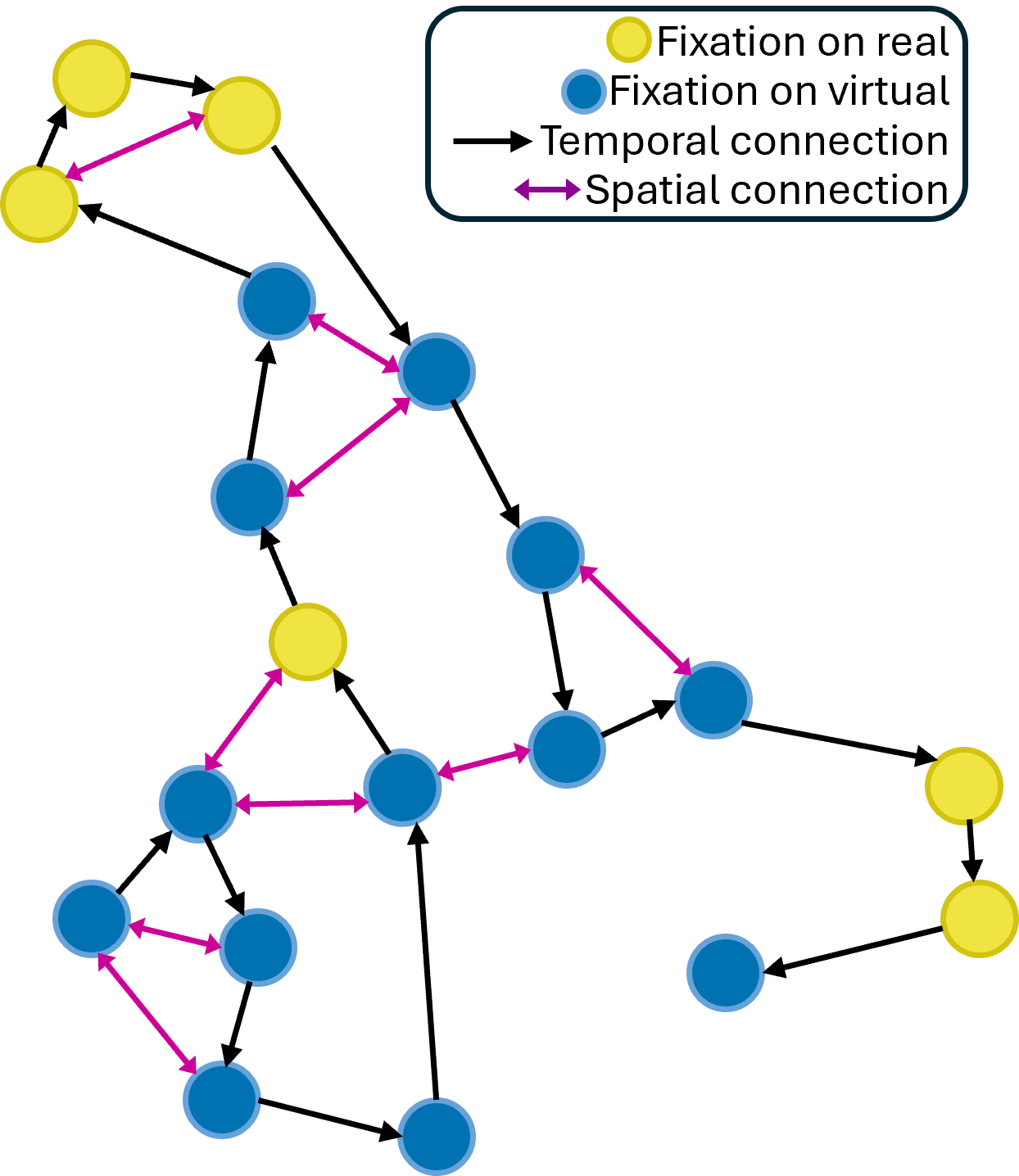}}
	\hspace{0.2mm}
	\subcaptionbox{FixGraphPool architecture}
	{\includegraphics[height=6.4cm]{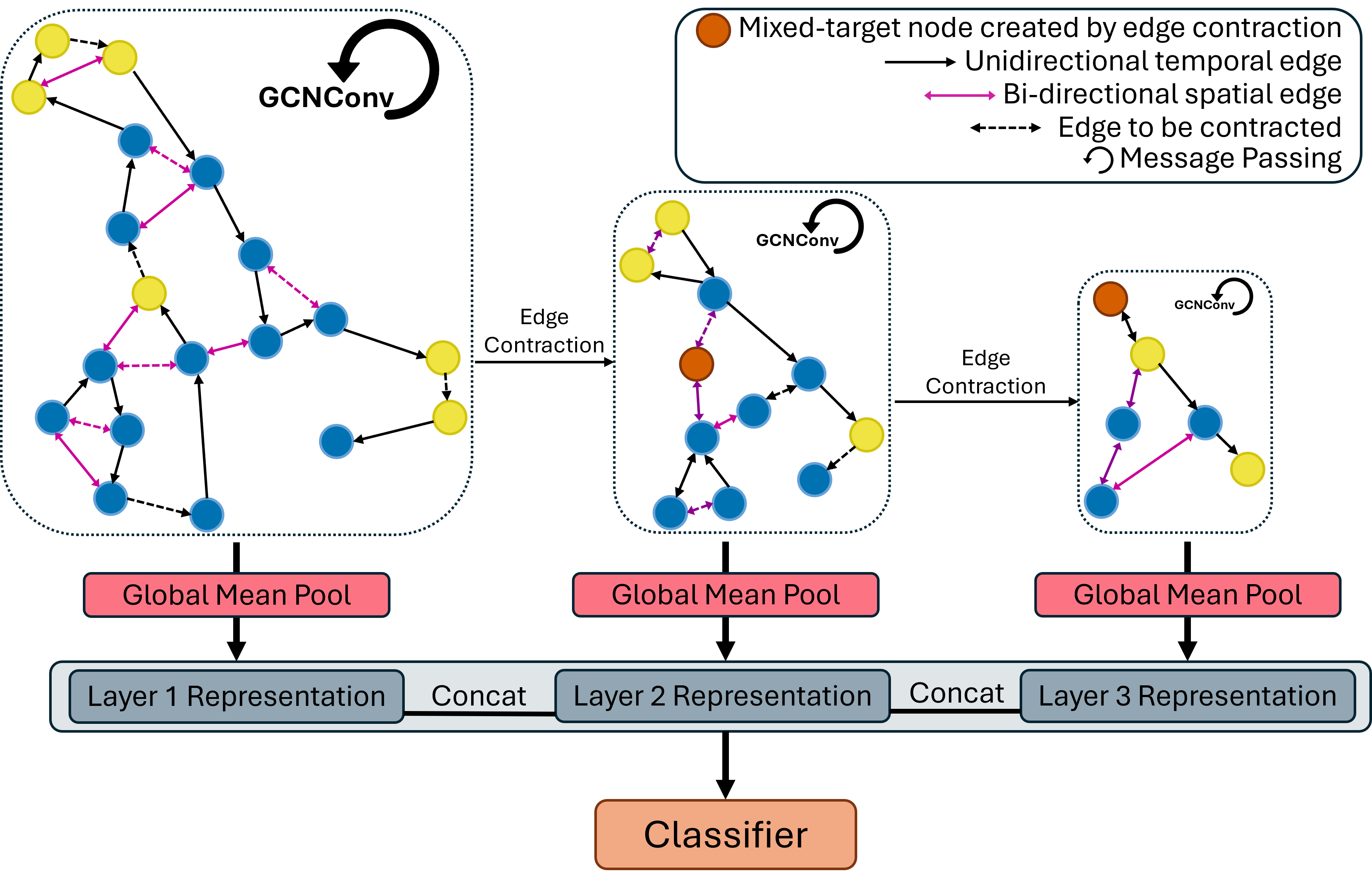}}
  \vspace{-8px}
	\caption{Illustration of GNN-based SA modeling. (a) Graph constructed from a window of gaze data, with fixation nodes colored based on target being virtual or real. (b) The proposed FixGraphPool architecture. Within each layer, GCN is first applied for message passing, followed by the computation of edge scores and then the pooling procedure by edge contraction. Dashed lines indicate edges that will be contracted to form a smaller new graph. Note that the color coding here is only for ease of visualization. Best viewed in color.}
    \vspace{-19px}
    \label{fig: gnn}
\end{figure*}

\vspace{-5px}
\section{FixGraphPool: A Graph Neural Network for Prediction of SA with ET Data}
\vspace{-4px}
\label{sec: gnn}
Beyond understanding the relationship between SA and ET metrics, we aim to develop a model capable of predicting SA based on ET data. This task presents three primary challenges:
\begin{itemize}[itemsep=0pt, topsep=-0.3pt, parsep=0.1pt, leftmargin=0.1in]
  \item \textit{Spatial and temporal encoding}: ET data is inherently temporal, and the spatial information of gaze targets encodes critical insights into visual processing strategies and attention distribution.
  \item \textit{ET in dynamic and unconstrained environments}: Our task, like other AR tasks requiring SA, involves dynamic environments with no constraints on head or eye movements. Unlike structured tasks such as reading, no specific gaze patterns are expected.
  \item \textit{User heterogeneity and data scarcity}: Users inherently employ different strategies for the task, and their ET data exhibit significant variability. Additionally, with 30 participants each performing two trials of no more than one minute, the amount of data available for training is limited.
\end{itemize}
\vspace{-1px}
In this section we propose \textit{FixGraphPool}, a model tailored to address these challenges using fixations extracted from raw data and relations between fixations to model user eye behaviors.

\vspace{-4px}
\subsection{GNN Preliminaries}
\vspace{-3px}
\noindent\textbf{Graphs} are defined as $\mathcal{G} = (\mathcal{V}, \mathcal{E})$, where $\mathcal{V}$ represents nodes and $\mathcal{E}$ denotes edges connecting node pairs, commonly used for modeling relations. 
\textbf{Graph neural networks (GNNs)} operate on graph-structured data typically through iterative message passing between nodes. Each node $v \in \mathcal{V}$ with feature vector $\mathbf{x}_v$ updates its representation $\mathbf{h}_v^{(l)}$ at layer $l$ by aggregating messages from its neighborhood $\mathcal{N}(v)$:
\begin{equation}
    \mathbf{h}_v^{(l)} = f^{(l)}\left(\mathbf{h}_v^{(l-1)}, \bigoplus_{u \in \mathcal{N}(v)} \phi^{(l)}\left(\mathbf{h}_v^{(l-1)}, \mathbf{h}_u^{(l-1)}, e_{vu}\right)\right)
\end{equation}
where $\bigoplus$ is a permutation-invariant aggregation function (e.g., sum, mean, max), $\phi$ computes pairwise messages, and $f$ updates the node's representation $\mathbf{h}_v^{(l)}$ at layer $l$ by combining aggregated information with its previous state $\mathbf{h}_v^{(l-1)}$. This allows GNNs to capture structural patterns and feature-based relationships. For \textbf{graph-level classification}, final node representations are aggregated (e.g., via mean, max, or attention pooling~\cite{li20242}) into a global graph representation $\mathbf{h}_\mathcal{G}$, which is then used for prediction.

Previous graph-based modeling of ET data, specifically GazeGraph~\cite{lan2020gazegraph}, typically represents raw 2D gaze points as nodes. In such models, edges encoding spatial-temporal relationships form an adjacency matrix, often processed with CNNs using edge weights derived from 2D distances. However, these approaches are constrained by their 2D input and omission of node-level features, limiting their ability to fully leverage the rich spatial-temporal dynamics in gaze behavior. To address these limitations, we propose a novel GNN designed to exploit these relationships in 3D gaze data.

\vspace{-5px}
\subsection{Task Formation}
\vspace{-3px}
Given the stability of ET metrics over time, as demonstrated in the previous section, we formulate the task as predicting whether a user will exhibit \textbf{\textit{poor}} SA based on small windows of ET data captured within the 21 seconds preceding the incident. Considering that the maximum duration of all extracted fixations is 6.12 seconds (note that this includes fixations identified using the ``loose'' definition in Sec.~\ref{sec: gaze-processing}), we set the window length to 7 seconds to ensure that each window contains at least one fixation. A sliding-window approach with a step size of 1 second is employed to extract the complete graph dataset from the B and V trials of all participants. The task is thus formalized as a binary graph classification problem, where the objective is to predict whether a given graph corresponds to a trial with \textit{poor} SA. We envision that such a model can be used to predict, with a 7s observation of user ET data, whether the user's SA is likely to be negatively affected, making it possible to provide timely feedback to the user and potentially mitigate such effects.

\vspace{-5px}
\subsection{Graph Construction}
\vspace{-3px}
Given the heterogeneity and limited size of our dataset, directly using raw gaze points as input to a model often results in severe overfitting. To address this, we leverage fixations, a well-established domain-specific feature in ET research, known to encapsulate shared information across subjects. For each window, we construct a graph where nodes represent fixations, and edges capture the spatial-temporal relationships between them.

\noindent\textbf{Nodes}. Each node represents a fixation extracted from the window data, whose features include (1) fixation direction represented in spherical coordinates, (2) cyclopean eye center position (in the marker's coordinate system), (3) fixation duration and (4) fixation target being virtual content or not. This results in a 7-dimentional feature vector for each node. \revise{Pupil dilation was discarded from node features due to its unreliable measurement during CPR, with approximately 15\% of the recorded pupil diameters registering at an implausibly low value of 1~mm~\cite{mathot2018pupillometry}, possibly due to the excessive movement during CPR. }

\noindent\textbf{Edges}. To represent relationships between fixations, we define two types of connections: \textit{temporal} and \textit{spatial}. \textit{Temporal connections} are modeled as directed edges linking temporally consecutive fixations. Unlike the GazeGraph approach, which employs bi-directional edges for all neighboring nodes, we retain the temporal edges as directed, emphasizing the sequential nature of temporal information flow. \textit{Spatial connections} are added between fixations not connected by temporal edges that are in close proximity. To determine proximity, we leverage task-specific information about the mannequin's height and project fixations onto the mannequin's plane to obtain two-dimensional coordinates. We then add bi-directional edges if the Euclidean distance between two fixations points is less than 6.5 cm---the radius of the larger ring in the compression depth visualizer, which we consider as the threshold for spatial closeness. For fixations off the mannequin, we assign fixed coordinates (0~m, 2~m) for virtual targets (in this case the timer text) and (2~m, 2~m) for other off-mannequin fixations, within the marker's coordinate system which is right-hand with the x-axis pointing to the waist of the mannequin. This ensures that such fixations are not connected to those on the mannequin, where fixation points span no more than 1 m both horizontally and vertically from the marker position. While this approach may result in non-virtual off-mannequin fixations being marked at the same location and thus connected, such cases are rare and typically encode a similar level of visual information, minimizing any adverse impact. An example of the constructed graph is shown in Fig.~\ref{fig: gnn}(a). It is important to note that the projection of gaze data onto a 2D plane is tailored to the specific requirements of this task and is not the sole method for determining spatial proximity. Alternative methodologies, such as projecting gaze directions and eye center positions into a spherical coordinate system and computing distances within that space, may be more appropriate for other task contexts. In this study, the 2D projection approach was selected due to the inherent limitations in head and eye tracking accuracy of Magic Leap 2.

\vspace{-1px}
\noindent\textbf{Edge features}. To capture the nuanced relationships between fixations, each edge is assigned a 3-dimensional feature vector. This vector encodes (1) the temporal difference between the two connected fixations, (2) the spatial distance between the fixation points (computed in the 2D space), and (3) a binary indicator specifying whether the target of the fixation at the edge's endpoint is virtual content. These features provide a comprehensive representation of both spatial and temporal dependencies in gaze behavior.

\vspace{-1px}
\noindent\textbf{Edge weights}. In addition to edge features, each edge is assigned a scalar weight to modulate the information flow between connected fixations. The weight is computed as a normalized exponential function of the spatial distance between the two fixations:
\vspace{-2px}
\begin{equation}
  W_{\mathbf{e}_{vu}} = \exp\left(-\frac{d_{vu}}{\sigma}\right)
  \vspace{-3px}
\end{equation}
where $d_{vu}$ represents the Euclidean distance (in 2D) between the two fixation points, and $\sigma$ is a normalization constant, set to 2 meters in this study. This weighting scheme ensures that spatially closer fixations exert a stronger influence during the graph message-passing process. Notably, the design of edge features and weights is independent of the edge type, ensuring general applicability.

\vspace{-8px}
\subsection{FixGraphPool Model}
\vspace{-4.5px}
In this section we present \textit{FixGraphPool}, an EdgePool-based~\cite{diehl2019edge} GNN model for SA prediction. EdgePool is a graph pooling method that iteratively contracts edges based on learned scores, effectively reducing the graph size while preserving important structural information. By dynamically selecting edges to pool, it enables hierarchical representation learning, making it suitable for tasks like graph classification. Our model shares the idea of pooling edges with EdgePool, but with two main improvements: \textit{multi-layer representation-fusion} and \textit{scoring edges with edge and node features together}. Diagram (b) in Fig.~\ref{fig: gnn} illustrates the overall model architecture. Below we describe the model in detail.

\vspace{-1px}
\noindent\textbf{Pooling layer}. In the pooling layer, a score is first computed for each edge based on the features of the two connected nodes \textit{and the edge features}. In the $l$th pooling layer, given an edge $\mathbf{e}_{vu}$ connecting nodes $v$ and $u$, we first compute the raw score $r$ as:
\vspace{-2px}
\begin{equation}
  r^{(l)}_{\mathbf{e}_{vu}} = \mathbf{W}^{(l)}\left(\mathbf{h}_v^{(l)}||\mathbf{h}_{\mathbf{e}_{vu}}^{(l)} || \mathbf{h}_u^{(l)} \right) + b^{(l)}
  \vspace{-3px}
\end{equation}
where $||$ denotes concatenation, $\mathbf{W}^{(l)}$ and $b^{(l)}$ are learnable parameters, and $\mathbf{h}_{\mathbf{e}_{vu}}^{(l)}$ is the edge feature vector. It is then normalized with all edges starting from node $v$ to obtain the pooling score $s^{(l)}_{\mathbf{e}_{vu}}$:
\begin{equation}
  s^{(l)}_{\mathbf{e}_{vu}} = \frac{\exp\left(r^{(l)}_{\mathbf{e}_{vu}}\right)}{\sum_{u \in \mathcal{N}(v)} \exp\left(r^{(l)}_{\mathbf{e}_{vu}}\right)} + C
  \vspace{-3px}
\end{equation}
where $C$ is a constant set as 0.5 following EdgePool's practice. Then starting from the edge with the highest score, we iteratively contract the edge and combine the connected nodes to form a new node $m$, whose feature vector is computed as:
\vspace{-2px}
\begin{equation}
  \mathbf{h}_{m}^{(l+1)} = (h_v^{(l)} + h_u^{(l)} + \mathbf{h}^{(l)}_{\mathbf{e}_{vu}}) \cdot s^{(l)}_{\mathbf{e}_{vu}} + 
  \begin{cases} 
    \mathbf{h}^{(l)}_{\mathbf{e}_{uv}} \cdot s^{(l)}_{\mathbf{e}_{uv}}, & \text{if } \mathbf{e}_{uv} \text{ exists} \\
    0, & \text{otherwise}
  \end{cases}
  \vspace{-3px}
\end{equation}
which is a combination of the two node features and the edge feature and the reverse edge feature if it exists, weighted by the pooling score. Different edges that connect two merged nodes will also be merged, whose new weight and feature are computed as:
\vspace{-2px}
\begin{equation}
  W^{(l+1)}_{\mathbf{e}_{mn}} = \sum_{i \in \{v, u\}, \\ j \in \{v', u'\}} W^{(l)}_{\mathbf{e}_{ij}} \quad \text{if } \mathbf{e}_{ij} \text{ exists}
\end{equation}

\begin{equation}
  \mathbf{h}^{(l+1)}_{\mathbf{e}_{mn}} = \sum_{i \in \{v, u\},  j \in \{v', u'\}} \mathbf{h}^{(l)}_{\mathbf{e}_{ij}} \quad \text{if } \mathbf{e}_{ij} \text{ exists}
  \vspace{-3px}
\end{equation}
where nodes $v, u$ are the two nodes being merged into $m$, and $v', u'$ are the two merged into $n$. Such edge contraction process is repeated iteratively, excluding the edges that have a newly-merged node as one of its endpoints. A new graph is formed after the pooling process with roughly half the number of nodes remaining.

\noindent\textbf{Message passing layer}. At the initial input and before each pooling layer, we apply a message-passing layer to update the node features. This is implemented with a standard GCN Layer~\cite{kipf2016semi}.

\noindent\textbf{Global pooling and classification}. After each pooling layer, we apply a global mean pooling layer to obtain a graph-level representation $\mathbf{h}_\mathcal{G}^{(l)}$ by averaging the node features. The final graph representation is obtained by concatenating the representations from all layers, which is then fed into a two-layer multilayer perceptron (MLP) for classification.

\vspace{-4px}
\subsection{Experiment Setup}
\vspace{-2px}
\textbf{Evaluation strategies.} We employed 5-fold cross-validation with participant-level separation between training and test sets to ensure only unseen users were included in the test set. \revise{To address label imbalance, we randomly oversampled (directly duplicating) users in the training set to balance the distribution (commonly applied on imbalanced data~\cite{chlopowiec2023counteracting,ilgin2023comparison}). As a result, the training set has 435 samples for both \textit{good} and \textit{poor} SA, while the test set maintains the original distribution with 75 \textit{poor} and 105 \textit{good} SA samples.}
\vspace{-0.5px}

\noindent\textbf{Metrics.} We report the mean and standard deviation of accuracy, F1 score, precision, and recall across all folds.

\vspace{-0.5px}
\noindent\textbf{Baselines.} We evaluated our model against several baselines: (1) logistic regression (LR), (2) support vector machine (SVM), (3) decision tree, (4) random forest, (5) AdaBoost with logistic regression (LR) as the base learner, and (6) AdaBoost with decision tree (DT) as the base learner. Additionally, we compare with PatchTSMixer~\cite{ekambaram2023tsmixer}, a SOTA time series classification model. For traditional ML models, we used the features described in Sec.~\ref{sec: gaze-metrics}, excluding MPD due to its negative impact on performance, likely caused by measurement inaccuracies. We also experimented with standard deviations of applicable metrics but found the performance to degrade. For PatchTSMixer, each 7-second window is treated as a time series, with gaze direction, eye center position, gaze target and the type of the gaze event that step belonged to (fixation, saccade or blink) as input features of each time step.
\vspace{-0.5px}

\noindent\textbf{Implementation details.} The model is implemented using PyTorch Geometric~\cite{fey2019fast}. Node and edge features are projected to 32 dimensions, with 3 pooling and 3 message-passing layers. Training is performed for 40 epochs using the AdamW optimizer (learning rate: 0.005, weight decay: 0.001), halving the learning rate every 5 epochs, with a batch size of 32. These hyperparameters were found through limited hyperparameter tuning. For PatchTSMixer, we used the official implementation, tuning hyperparameters extensively. The best configuration includes 4 layers, a patch length of 20, and 32 hidden units, trained for 10 epochs with a batch size of 32, a learning rate of 0.001, and a CycleLR scheduler. Classical ML models are implemented with Scikit-learn~\cite{scikit-learn}, with hyperparameters optimized via grid search using an inner 5-fold cross-validation on the training set. We chose the feature set that yielded the best performance for each classical ML model.

\noindent\textbf{Ablation study.} We conducted ablation study on a PatchTSMixer variant excluding gaze event input and several FixGraphPool variants: (1) using \textit{b}i-\textit{d}irectional \textit{t}emporal \textit{e}dges (BDTE), (2) not using \textit{m}ulti-\textit{l}ayer graph \textit{r}epresentation\textit{ f}usion (MLRF) (3) excluding \textit{e}dge \textit{f}eatures in \textit{p}ooling \textit{s}core computation (EFPS).

\vspace{-6px}
\subsection{Evaluation of FixGraphPool}
\vspace{-3px}
Table~\ref{tab: ml-results} summarizes the results. FixGraphPool achieved the best performance, with 83.0\% accuracy and an F1 score of 80.1\%, outperforming all baselines. PatchTSMixer suffered greatly from overfitting, achieving only 74.5\% accuracy and 67.5\% F1 score, even when excluding gaze event input. This highlights the challenges of modeling dynamic 3D ET data without domain knowledge---with the heterogeneity of user behaviors, the SOTA model was outperformed by classic ML models as simple as LR (76.4\% accuracy and 70.2\% F1 score). FixGraphPool mitigated user behavior heterogeneity by utilizing fixations and encoding spatial-temporal relationships, resulting in significantly improved performance.

Ablation studies confirmed the importance of unidirectional temporal edges, multi-layer representation fusion, and edge features in pooling scores. While excluding edge features was shown to slightly improve precision, it also increased false negatives, as reflected in lower recall. These results further justified the design choices in the graph construction and modeling process.

\begin{table}[h]
  \centering
  \vspace{-8px}
  \caption{Model performance for \textit{poor} SA prediction.}
  \vspace{-8px}
  \label{tab: ml-results}
  \resizebox{\columnwidth}{!}{
  \begin{threeparttable}
  \begin{tabular}{@{}ccccc@{}}
  Model  & Accuracy (\%) & F1 (\%)& Precision (\%)& Recall(\%)\\ \midrule
  LR & 76.4$\pm$4.7 & 70.2$\pm$10.1 & 76.4$\pm$12.0 & 69.9$\pm$21.8 \\
  SVM & 67.0$\pm$4.6 & 58.1$\pm$5.7 & 63.7$\pm$10.1 & 54.7$\pm$9.1 \\
  Decision Tree & 64.9$\pm$8.5 & 60.3$\pm$7.6 & 59.3$\pm$11.6 & 62.6$\pm$9.3 \\
  Random Forest & 70.3$\pm$3.7& 59.9$\pm$10.6 & 68.9$\pm$4.2 & 55.9$\pm$20.5 \\
  AdaBoost LR & 73.6$\pm$3.8 & 67.0$\pm$7.7 & 70.8$\pm$5.1 & 65.4$\pm$15.8 \\
  AdaBoost DT & 67.9$\pm$5.8 & 55.5$\pm$15.2 & 64.5$\pm$3.8 & 52.1$\pm$24.4 \\
  \midrule
  \textbf{PatchTSMixer} & 71.0$\pm$5.0 & 60.8$\pm$9.3 & 67.8$\pm$8.5 & 55.4$\pm$16.8 \\
  $-$gaze event & 74.5$\pm$7.4 & 67.5$\pm$12.6 & 70.0$\pm$9.7 & 66.8$\pm$17.9 \\
  \midrule
  \textbf{FixGraphPool}& \textbf{\underline{83.0$\pm$5.5}} & \textbf{80.1$\pm$5.2}& \underline{79.2$\pm$12.7} & \textbf{84.5$\pm$14.4} \\
  $+$BDTE & \underline{81.7$\pm$5.1} & \underline{77.8$\pm$7.4} & 77.0$\pm$9.9 & \underline{82.3$\pm$17.5} \\
  $-$MLRF & 76.6$\pm$4.5 & 71.9$\pm$7.5 & 70.6$\pm$7.8 & 76.0$\pm$17.5 \\
  $-$EFPS& \underline{81.7$\pm$8.0} & 75.4$\pm$12.9 & \textbf{79.7$\pm$5.1} & 72.9$\pm$18.9 \\
      \bottomrule
  \end{tabular}
  \begin{tablenotes}
      \footnotesize
      \item[1] The best results are in \textbf{bold} and the second best are \underline{underlined}.
      \end{tablenotes}
      \vspace{-10px}
  \end{threeparttable}
  }
  \vspace{-8px}
\end{table}

\vspace{-2px}
\section{Discussion And Future Work}
\label{sec: discussion}
\vspace{-3px}
In this work, we have identified correlations between overfocus on virtual elements, reduced attention shifting, and poor SA in AR-guided CPR. Based on this, we proposed FixGraphPool, which uses gaze targets, fixation duration, and saccade features to predict SA. While this work establishes a foundation for understanding ET metrics and SA, further exploration is needed.

\noindent\textbf{Impact of interface design.} This study's findings are rooted in a specific AR interface design, encompassing particular virtual elements, color coding, and visualization placements. While this design was informed by pilot testing and expert input, its applicability to other AR interfaces warrants consideration. For instance, the inclusion of elements such as the timer and compression statistics text may have imposed a high visual load on some participants. The adopted color scheme, originally developed for force visualization, might not be universally optimal, especially given that red and green are perceived more rapidly than blue and yellow~\cite{woodward2023designing}, and its effectiveness can depend on background colors. Furthermore, the placement of visualizations can influence user attention due to the established leftward visual bias in adults~\cite{bowers1980pseudoneglect,woodward2023designing}. The drumbeat audio could also have affected SA~\cite{davis2007effects}. Crucially, both our SA labeling methodology and the FixGraphPool model were designed to be agnostic to these specific interface characteristics. Future research will explore the impact of varied interface designs on SA and ET metrics, aiming to leverage these insights for developing SA-aware systems that can utilize FixGraphPool predictions to mitigate SA impairments through adjustment to interface.

\noindent\textbf{Relationship between self-evaluations, SA, and ET metrics.} The observed negative correlation between self-reported attention-shifting ability (ATTCS) and both SA and saccade metrics may seem counterintuitive. Interestingly, this finding resonates with multitasking research, which suggests that individuals frequently overestimate their multitasking prowess, and that those who multitask most often may possess lower working memory capacity and executive control, despite greater confidence in their abilities~\cite{kiesel2022handbook,sanbonmatsu2013multi}. While one might expect this correlation to translate into impaired CPR performance among participants with higher ATTCS scores, such an effect was not evident in our study. This absence could be due to the limited sample size or the lack of stringent CPR quality enforcement. Future investigations should explore this relationship more deeply, employing controlled experimental designs and larger participant groups. These results also underscore potential AR-induced hazards for users inclined to multitask, emphasizing the importance of designing AR systems that account for these individual differences to prevent adverse outcomes.

\noindent\textbf{Model generalizability.} FixGraphPool was designed to address the data scarcity and user heterogeneity inherent in SA evaluation. Scarcity is a consequence of the task design, where only the brief window before an incident can be reliably labeled. Heterogeneity stems from unmeasured user-specific factors, such as the use of peripheral vision, proficiency, or physical traits. For example, some taller participants maintained SA with minimal gaze shifts, possibly due to a wider peripheral view within the headset's field of view constraint. To counteract these challenges and prevent overfitting, FixGraphPool leverages domain knowledge by modeling gaze events (fixations) and their spatiotemporal relationships, combined with a pooling mechanism. While promising for tasks with similar data constraints, the model's generalizability must be validated across different applications, user groups, and data scales. Future work will explore the application of similar graph-based models to other AR contexts, such as activity recognition~\cite{lohr2023gazebasevr} or reading analysis~\cite{yang2025reading}, with larger and more diverse datasets. 

\vspace{-7px}
\section{Conclusion}
\vspace{-3px}
In this work, we investigated SA in AR-guided CPR tasks, focusing on the modeling of SA using ET data. Through a user study with realistic incidents, we identified key ET metrics correlated with SA and proposed FixGraphPool, a GNN model leveraging gaze events and spatial-temporal relation in between for SA prediction. Our model outperformed traditional machine learning and SOTA time-series models, achieving 83.0\% accuracy. These findings highlight the potential of ET data for SA modeling and provide insights into designing AR systems that balance task guidance with environmental awareness, ensuring user safety in critical scenarios.
\label{sec: conclusion}

\vspace{-7px}
\acknowledgments{
  \vspace{-3px}
  We thank Cliff Merritt, Bradi Granger and Anna Mall for sharing their insights on the study and app design; we thank Prof.~David Carlson and Dr.~Amy McDonnell for helpful discussions regarding the work and all participants for contributing to the study. This work was supported in part by NSF grants CSR-2312760, CNS-2112562, and IIS-2231975, NSF CAREER Award IIS-2046072, NSF NAIAD Award 2332744, a CISCO Research Award, a Meta Research Award, Defense Advanced Research Projects Agency Young Faculty Award HR0011-24-1-0001, and the Army Research Laboratory under Cooperative Agreement Number W911NF-23-2-0224. The views and conclusions contained in this document are those of the authors and should not be interpreted as representing the official policies, either expressed or implied, of the Defense Advanced Research Projects Agency, the Army Research Laboratory, or the U.S. Government. This paper has been approved for public release; distribution is unlimited. No official endorsement should be inferred. The U.S. Government is authorized to reproduce and distribute reprints for Government purposes notwithstanding any copyright notation herein.}

\bibliographystyle{IEEEtran}
\bibliography{custom}
\end{document}